\pdfoutput=1

\documentclass[11pt]{article}

\usepackage[]{acl}

\usepackage{times}
\usepackage{latexsym}
\usepackage{tabularx}
\usepackage{xcolor}

\usepackage[T1]{fontenc}

\usepackage[utf8]{inputenc}

\usepackage{booktabs}

\usepackage{microtype}

\usepackage{hyperref}

\usepackage{graphicx}
\usepackage{amsmath}
\usepackage{amssymb}
\usepackage{dblfloatfix}
\usepackage{multirow}
\usepackage{array, makecell} %

\usepackage{algorithm}
\usepackage[noend]{algorithmic}
\newcommand{\algorithmicinput}{\textbf{input}}
\newcommand{\INPUT}{\item[\algorithmicinput]}

\DeclareMathOperator*{\argmax}{argmax}

\title{Efficient Unsupervised Sentence Compression by Fine-tuning Transformers with Reinforcement Learning}

\author{Demian Gholipour Ghalandari$^{1,2}$, Chris Hokamp$^1$, \\
\textbf{Georgiana Ifrim}$^2$\\
$^1$Aylien Ltd., Dublin, Ireland \\
$^2$Insight Centre for Data Analytics, University College Dublin, Ireland \\
$^1$\texttt{\{first-name\}@aylien.com} \\
\texttt{georgiana.ifrim@ucd.ie}}

\begin{document}
\maketitle
\begin{abstract}
\end{abstract}
Sentence compression reduces the length of text by removing non-essential content while preserving important facts and grammaticality. Unsupervised objective driven methods for sentence compression can be used to create customized models without the need for ground-truth training data, while allowing flexibility in the objective function(s) that are used for learning and inference. Recent unsupervised sentence compression approaches use custom objectives to guide discrete search; however, guided search is expensive at inference time. In this work, we explore the use of reinforcement learning to train effective sentence compression models that are also fast when generating predictions. In particular, we cast the task as binary sequence labelling and fine-tune a pre-trained transformer using a simple policy gradient approach. Our approach outperforms other unsupervised models while also being more efficient at inference time.

\section{Introduction}
\label{sec:introduction}

In general, the information content of text is correlated with its length. However, for a given text, a shorter version may still convey the essential information while preserving grammaticality \citep{Siddharthan2014ASO}. The definition of \textit{essential} can change depending on the downstream application, thus models for text compression must be able to adapt based on information about the downstream task.

Sentence compression models have been used as sub-modules of text and speech summarization \citep{banerjee2015multi, shang-etal-2018-unsupervised}, for headline generation \cite{dorr-etal-2003-hedge}, subtitle generation \citep{vandeghinste-pan-2004-sentence}, and summarizing emails \citep{zajic2008single}. Potential applications also include snippet generation and highlighting for social media, blog posts or search results.

\begin{figure}[ht!]
\centering
\includegraphics[width=1.0\columnwidth]{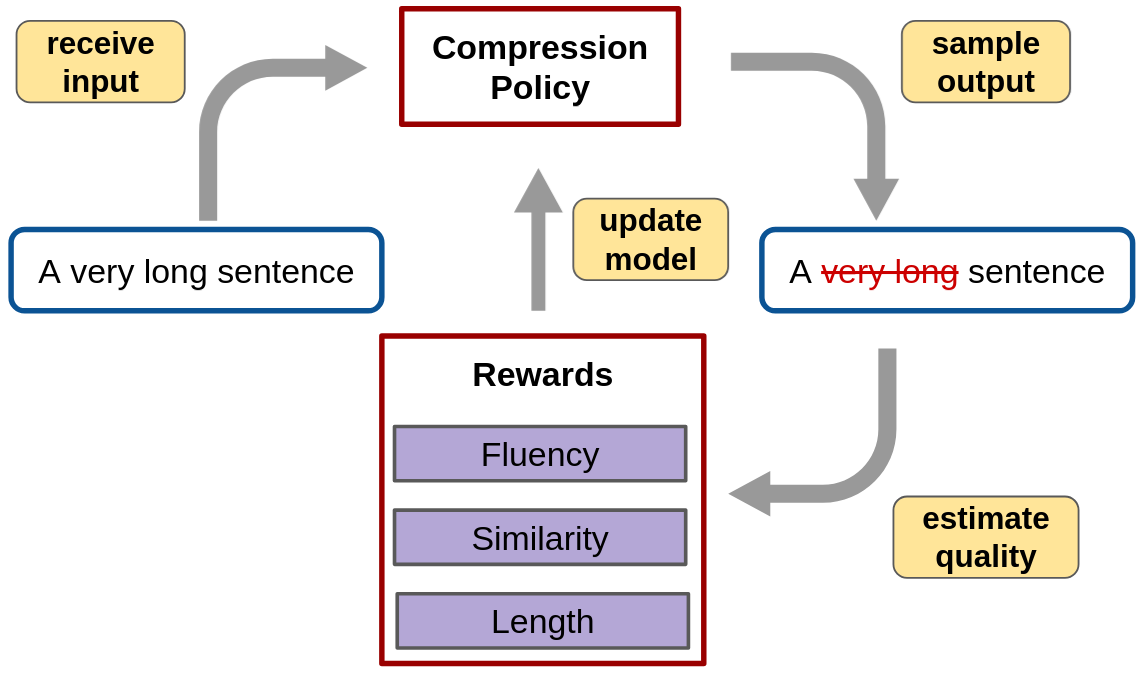}
\caption{Reinforcement learning framework for sentence compression.}
\label{fig:framework}         
\end{figure}

Given a particular text compression task, relevant evaluation metrics and auxiliary models of compression quality may not be straightforward to formulate as  well-behaved differentiable objectives that can be used with standard backpropagation. In addition, ground-truth examples may be difficult to obtain because the annotation task is difficult to fully specify, and metrics which capture different facets of compression quality, such as fluency and optimal sentence length, may be negatively correlated. Even in the case where ground-truth examples are available, they are likely to represent only a subset of the possible outputs, so there is a  risk of over-fitting or biasing models when relying solely upon a small amount of gold training data for optimization. 

Recent unsupervised sentence compression approaches leverage powerful neural language models to directly optimize objectives such as fluency and faithfulness of compressed sentences, using discrete search strategies, without relying on ground-truth examples \citep{niu2019deleter, zhou2019simple, schumann2020discrete}. However, these search-based methods are very inefficient at inference-time because the search must navigate through a large candidate space while recomputing expensive reward functions. 



\begin{figure}[!t]
\centering
\includegraphics[width=0.8\columnwidth]{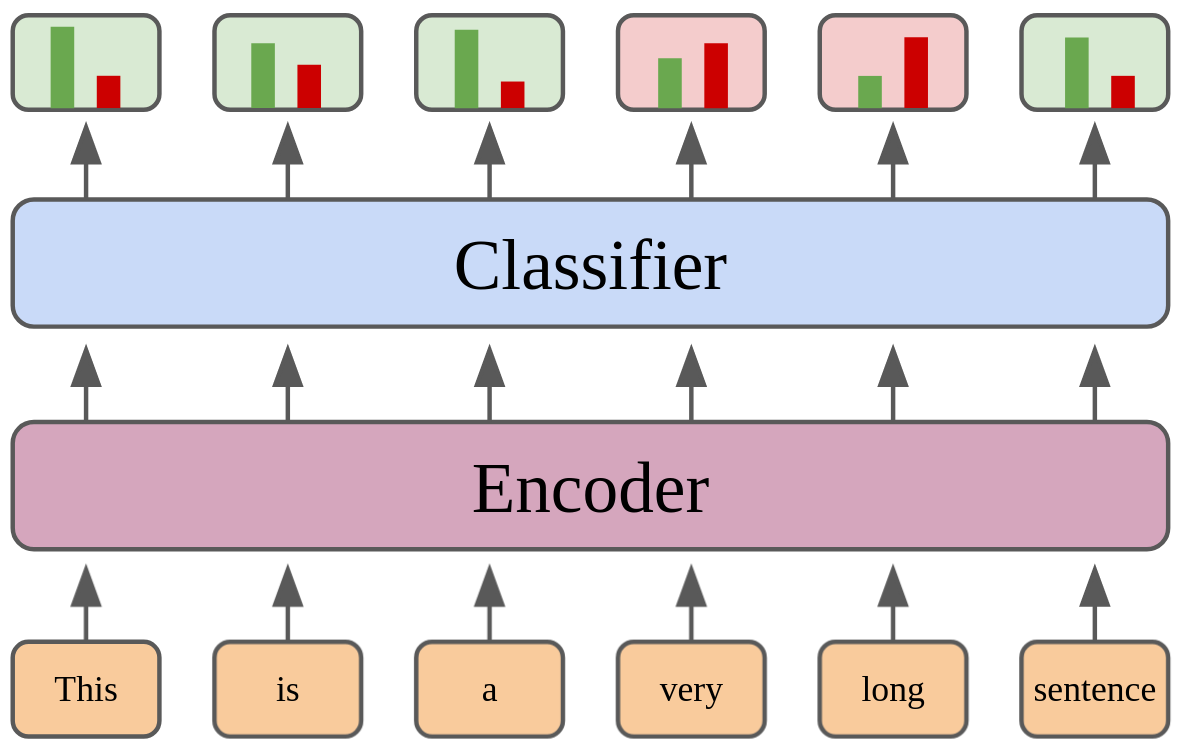}
\caption{Model architecture for compressing sentences.}
\label{fig:architecture}         
\end{figure}

To allow for flexible reward specification, while also enabling efficient inference, we design a simple and effective reinforcement learning (RL) setup: our model is initialized as an unsupervised pre-trained language model with an untrained binary classification head (see Figure \ref{fig:architecture}), and the sentence compression task is framed as sequence labeling, with optimization via policy gradient using a suite of reward functions. Sentences are compressed in an instantaneous, one-step fashion, similar to modern part-of-speech tagging or named entity recognition models. This approach simplifies the learning setup while also allowing for high throughput.

According to quantitative evaluation on several summarization benchmarks, our approach shows similar or superior performance compared to search-based methods, while also being much faster at inference time.

Our approach to unsupervised extractive sentence compression has the following benefits:
\begin{itemize}
    \vspace{-0.2cm}
    \item \textbf{Unsupervised:} No labelled examples are required.
    \vspace{-0.2cm}
    \item \textbf{Fast inference:} At test time, the model only performs one-step sequence labeling.
    \vspace{-0.2cm}
    \item \textbf{Configurable:} Rewards can be tailored to specific use cases.
\end{itemize}

\vspace{-0.2cm}
We review related work in Section~\ref{sec:related-work}. Section~\ref{sec:task} formalizes the task. Section~\ref{sec:method} gives a detailed description of the model and reward functions. Section~\ref{sec:experiments} presents experimental results, and Sections \ref{sec:analysis} and \ref{sec:discussion} provide analysis and discussion of our findings. 

\section{Related Work}
\label{sec:related-work}

\subsubsection*{Unsupervised Sentence Compression}



Early work on sentence compression casts the task as an optimization problem under linguistically motivated constraints \citep{HoriF04-speech-summ,clarke-lapata-2006-models,Clarke:Lapata:2008-ilp}. The objectives to be optimized include n-gram language model scores and frequency-based word relevance measures. Constraints are designed to ensure the grammaticality of compressions.

Some recent work follows the discrete optimization paradigm while leveraging powerful models as objective functions in place of hand-crafted constraints, while exploring different strategies for heuristic search: \citet{zhou2019simple} use beam search to optimize a fluency and a similarity objective. \citet{schumann2020discrete} use a greedy hill-climbing search to optimize fluency and similarity objectives. \citet{niu2019deleter} use a greedy search with a look-ahead mechanism, only optimizing fluency. All of these recent approaches use large neural language models to estimate fluency. While the approach presented in whis work does not involve discrete search, we consider it complementary and orthogonal to our RL-based approach (see Section \ref{sec:discussion} for more discussion).

Another commonly proposed unsupervised framework is to use autoencoders and reconstruction objectives \citep{miao-blunsom-2016-language, fevry-phang-2018-unsupervised, malireddy-etal-2020-scar}. These approaches are based on the assumption that a good sentence compression is one from which the original sentence can be inferred.

\citet{wang2018sentence} is an example of prior work using reinforcement learning for unsupervised sentence compression. They use a Deep Q-Network to optimize a reward incorporating n-gram language model probabilities and grammatical constraints. This model repeatedly deletes a token until it terminates, as opposed to our one-step approach. \citet{zhao2018language} also use RL to optimize a syntax-focused language model score. However, their policy is initialized with a supervised sentence compression model, whereas ours is fully unsupervised.

\subsubsection*{Reinforcement Learning for Summarization}

Reinforcement learning has become popular in the wider field of text summarization, finding applications in both  extractive and abstractive sub-tasks. One use case of RL is in supervised scenarios, where rewards are computed based on ground-truth examples, e.g., ROUGE scores, to overcome issues with cross-entropy losses \citep{paulus2017deep, narayan-etal-2018-ranking, dong-etal-2018-banditsum}. 
\textsc{BanditSum} \citep{dong-etal-2018-banditsum} in particular has a very similar RL setup to ours: they train in one-step episodes where a policy predicts extractive labels and immediately receives a reward. \citet{scialom-etal-2019-answers} augment a ROUGE-based reward with a reward based on question answering.
\citet{bohm-etal-2019-better} and \citet{stiennon2020learning} learn reward functions from human quality ratings of summaries.
Similar to our unsupervised approach, \citet{laban2020summary} use RL for unsupervised abstractive summarization, optimizing reward functions representing fluency, coverage under a length constraint, and also use a policy gradient approach.

\section{Task}
\label{sec:task}
We focus on the specific task of summarizing a sentence by extracting a subset of its tokens in their original order. Given an input sentence $x$ consisting of $n$ tokens $x = (x_0, x_1, ..., x_{n})$, we aim to produce a sequence of binary labels $y = (y_0, y_1, ..., y_n) \in \{0, 1\}^{n}$, where each label indicates whether the corresponding input token should be included in the compressed version of a sentence.

We further assume an objective function, or \textit{reward} function $R(x, y)$ that measures how well applying the labels $y$ summarizes the original sentence $x$. For a particular $x$, the goal is to find $\argmax_{y} R(x, y)$, without access to any ground-truth examples.

In general, there are $2^n$ possibilities to shorten a sentence in this task. A fixed summary length $L$ would reduce this to $n \choose L$ possibilities, peaking at $L = \frac{n}{2}$ (for even $n$). We do not constrain our approach to a fixed length, but we compare it to search-based techniques that are constrained to the $n \choose L$ search space.

\section{Method}
\label{sec:method}

\subsection{Training Procedure}

We train a policy $\pi_{\theta}$ with parameters $\theta$ to produce binary labels. 
Given an input $x$, the policy $\pi_{\theta}$ predicts a binary keep/discard probability distribution for each token index in $x$.  We use the notation $\pi_{\theta}(*|x)$ to refer to the collection of these distributions for all tokens in $x$. We obtain the probability $\pi_{\theta}(y|x)$ of a label sequence $y$ given input sequence $x$ as follows:


\begin{equation}
\pi_{\theta}(y|x) = \prod_{i} \pi_{\theta}(y_i|x),
\end{equation}

\noindent where $ \pi_{\theta}(y_i|x)$ is the probability of a token $x_i$ being included if $y_i=1$ or excluded if $y_i=0$. To compress a sentence using $\pi_{\theta}$, we select the higher scoring label for each token:
\begin{equation}
    y^a = \{ \argmax_{y_i} \pi_{\theta}(y_i|x)\ \text{ for } y_i \in y \}. 
\end{equation}

We train our model using a policy gradient technique \citep{DBLP:conf/nips/SuttonMSM99}. Unlike typical sequential reinforcement learning scenarios, our $\pi_{\theta}$ only performs one action for a given input, receiving the corresponding reward immediately, without transitioning through other intermediate states. Therefore, our setup is similar to a contextual multi-armed bandit problem \cite{NIPS2007_4b04a686}, where each "arm" corresponds to a particular label sequence $y = (y_0, y_1, ..., y_n) \in \{0, 1\}^{n}$. However, in our scenario, the policy is generally allowed to access rewards for multiple possible actions via sampling, which is different from typical bandit settings where only one $ (action, reward) $ pair is available for each episode.

The training objective is to maximize the expected reward assigned to a predicted label sequence $y$ for a given input $x$, computed by the reward function $R$:

\begin{equation}
    J(\theta) = \mathbb{E} [ R(x, y) ]
\end{equation}

The policy gradient theorem states that the gradient of this expectation can be expressed as follows \citep{DBLP:conf/nips/SuttonMSM99}:

\begin{equation}
    \nabla_{\theta} J(\theta) =  \nabla_{\theta} \mathbb{E} [ R(x, y) \text{ log } \pi_{\theta} (y|x) ]
\end{equation} 
Since the above expectation is intractable for a large dataset and the corresponding action space, this gradient is estimated by sampling:

\begin{equation}
    \nabla_{\theta} J(\theta) =  \nabla_{\theta} r^s \text{ log } \pi_{\theta} (y^s|x),
\end{equation} 

\noindent where $y^s \sim \pi_{\theta}(*|x)$ is a sample from the current policy at a given step, consisting of binary token labels $y^s = (y{^{s}}_0, y{^{s}}_1, ..., y{^{s}}_n)$, and $r^s = R(x, y^s)$.

As is commonly done when using policy gradients, we subtract a baseline from the reward for variance reduction. We instantiate the baseline as $r^a = R(x, y^{a})$, the reward given to the the most likely label sequence $y^a$ according to the current policy. The gradient becomes:
\begin{equation}
    \nabla_{\theta} J(\theta) =  \nabla_{\theta} (r^s - r^a) \text{ log } \pi_{\theta} (y^s|x)
\end{equation} 

\noindent Accordingly, we train our model by minimizing the following loss function:

\begin{equation}
    \ell_{\theta} = (r^a - r^s) \textrm{ log } \pi_{\theta}(y^s|x).
\end{equation}

Using the baseline $r^a$ allows the intuitive interpretation that a sample $y^s$ is encouraged if its reward is higher than the current policy's prediction, i.e., when factor $(r^a - r^s)$ is negative, and discouraged otherwise.

\subsection*{Best-of-$k$ Sampling}
Prior work with a similar application of policy gradient \cite{dong-etal-2018-banditsum, laban-etal-2021-keep} observed an advantage in sampling $k$ times and taking the average loss over all samples  rather than using a single sample. However, in our experiments, we observe that only using the sample with the \textit{maximum} reward from a large number of samples works significantly better than taking the average or only sampling once. A large $k$ improves the discovery of high-quality compressions -- if we only use a single sample or a very small $k$, we observe a higher tendency of models to converge on simple behaviors with low reward improvements, such as only extracting the first-$L$ tokens of a sentence. The choice of $k$ controls a trade-off: with a higher $k$, we spend more time computing the rewards of samples and less on model updates, given a limited wall-time constraint for training. We determine $k$ in an unsupervised manner using a validation set (details in Section \ref{subsec:model-dev}).

\subsection{Model Architecture}
$\pi_{\theta}$ is initialized as a transformer encoder model with a linear classification head. In particular, we use the 6-layer DistilRoBERTa model \citep{Sanh2019DistilBERTAD} due to its efficiency and smaller size compared to other BERT-like models, while retaining good results on the GLUE benchmark\footnote{\url{https://huggingface.co/distilroberta-base}}. During training, the whole model is fine-tuned. For each token in the input, our model will determine whether it should be kept or filtered. Figure \ref{fig:architecture} visualizes the design. This architecture produces summaries in an instantaneous, non-autoregressive fashion, allowing for fast prediction (see Section \ref{subsec:running-times}).

\subsection{Reward Functions}
\label{subsec:reward-functions}

We do not have direct access to ground-truth training data in our setup, so we consider a suite of reward functions that may correlate with different aspects of sentence compression quality.

\subsubsection*{Fluency}
This reward function is intended to ensure grammatically correct and well-written sentences. We use a masked language model (LM) to estimate the fluency of a compressed sentence. In particular, we compute fluency as the average logit of a token $y_i$ in the compressed sentence $y$. We do this without masking $y_i$ to reduce the running time during training, as masking would require to re-encode the sentence for each token. Based on our experiments, this simplification still produces good estimates of fluency.

\begin{equation}
    R_{fluency}(y) = \frac{1}{|y|} \sum_{i=1} \text{ LM}(y_i|y)
\end{equation}

We normalize $R_{fluency}$ by dividing it by an empirically set constant, to keep its values in a similar range compared to the other rewards. The constant is an observed minimum value from a sample dataset. We argue that a masked language model is more appropriate in our setup compared to a left-to-right (causal) language model -- when predicting or sampling a compressed sentence during training, the sentence is treated as a finished rather than an intermediate output, which is not captured by the auto-regressive inference of causal LMs. We confirm the advantage of a masked LM over a left-to-right LM in a comparison on a development set (Appendix \ref{appendix:masked-vs-causal}).

We note the precedent for using language models to measure fluency: \citet{zhou2019simple} and \citet{schumann2020discrete} use language models trained on a summarization target domain, e.g., headlines. \citet{laban2020summary} uses a generic causal language model to estimate fluency. \citet{niu2019deleter} use a masked language model to score candidate compressions.

\subsubsection*{Similarity-to-Source}
The similarity reward is intended to preserve the meaning of the source sentence in the compressed sentence. We experiment with several options to compute similarity, all using models from the \texttt{sentence-transformers} library\footnote{\url{https://www.sbert.net/}} \citep{reimers-gurevych-2019-sentence}: 
\begin{itemize}
    \itemsep0em
    \item \textbf{Bi-Encoder Similarity}: A sentence encoder $f$ separately computes embeddings for the source and the predicted summary. We calculate the cosine similarity between both embeddings: $R_{sim}(x, y) = cos(f(x), f(y))$
    \item \textbf{Cross-Encoder Similarity}:
    Output of a cross-encoder model $f_{sim}$ measuring the semantic textual similarity between both sentences: $R_{sim}(x, y) = f_{sim}(x, y)$
    \item \textbf{Cross-Encoder NLI:}
    We also test a natural language inference (NLI) model $f_{nli}$ to estimate how well a compressed sentence retains source information. The intuition is that the source should \textit{imply} information in the output:
    $R_{nli}(x, y) = f_{nli}(y|x)$
    
\end{itemize}

\noindent Based on experiments on a development dataset, the bi-encoder similarity performs best in our setup.

\subsubsection*{Length and Compression Ratio}
Because our model is non-sequential, we cannot easily employ a hard constraint to control the length of compressed sentences. Instead, we impose a soft length control using Gaussian reward functions. In particular, we either use a reward function for the length (token count) in a compressed sentence $R_{len}$, or one for the compression ratio between the source and prediction, in terms of token counts, $R_{cr}$. We choose one of these two depending on whether a consistent length or a consistent ratio is desired, which differs for different evaluation datasets. We set the distribution means of both rewards as the desired values for word count and compression ratio. We set the standard deviations as the mean times a factor $s$ which we set to 0.4 for both reward functions (Equations \ref{eq:len-reward}, \ref{eq:cr-reward}): 

\begin{equation}
    \label{eq:len-reward}
    R_{len} = \mathcal{N}(\mu_{L},\, (s \times \mu_{L})^{2}),
\end{equation}
\begin{equation}
    \label{eq:cr-reward}
    R_{cr} = \mathcal{N}(\mu_{cr},\,(s \times \mu_{cr})^{2}).
\end{equation}

\subsubsection*{Reward Aggregation}
The final reward function is an average of the reward functions $R_{fluency}$, $R_{sim}$, combined with either $R_{len}$ or $R_{cr}$:
\begin{equation}
    r_{total}(x, y) = \frac{1}{3} \sum_{i}R_i(x, y).
\end{equation}

\vspace{-0.2cm}
In practice, when the downstream task is known, reward functions may be designed and calibrated based upon insights and domain expertise, e.g., an optimal summary length for a specific application or different language models corresponding to different summary styles. In this work, we only use publicly available and commonly-used off-the-shelf models to construct reward functions. 



\section{Experiments}

\label{sec:experiments}
This section presents a detailed analysis and evaluation results for our proposed model. We name our model \textbf{\textsc{SCRL} (Sentence Compression with Reinforcement Learning)}. We make all code, model outputs and data available\footnote{\url{https://github.com/complementizer/rl-sentence-compression}}.

\subsection{Datasets}

\subsubsection{Training Datasets}
We use two datasets for training: Newsroom \citep{grusky-etal-2018-newsroom} and Gigaword \citep{Rush_2015}. For Newsroom, we extract the first three sentences from each article, only keeping sentences with a number of tokens between 15 and 60. Newsroom was chosen due to the large size and a variety of un-preprocessed news articles from different sources. Ground-truth summaries are not included in the training data, thus the two datasets are treated as large unlabeled text collections. We train a model for short headline-like summaries on Gigaword to evaluate it on the Gigaword test set, which comes in a specific preprocessed format\footnote{Lowercased, pre-tokenized, rare words and digits replaced with special tokens.}. Training on Gigaword allows to expose the model to the same preprocessing, for a fair evaluation.

\subsubsection{Development Dataset}
We constructed a small labelled validation dataset for model development: we automatically identified sentence-summary pairs in Newsroom, also including title-summary pairs, by extracting cases where the tokenized summary is contained in a tokenized sentence, with preserved order. We manually filter a subset of these examples based on grammaticality and informativeness and obtain 280 examples. This dataset was only used during initial development to compare the different reward function variants discussed in Section~\ref{subsec:reward-functions}.

\subsubsection{Evaluation Datasets}
The evaluation includes five test sets -- key statistics are listed in Table \ref{tab:datasets}. $L_{src}$, $L_{tgt}$ are the token counts in source and target sentences and $cr = L_{tgt} / L_{src}$ is the compression ratio. Following \citet{schumann2020discrete}, we compare our models on Gigaword against baselines of comparable length brackets using ROUGE F1-scores\footnote{We only consider lengths similar to the ground-truth, i.e. 8-10 tokens.}. For DUC2004 (Task 1), following prior work, we truncate model outputs to 75 characters and compute ROUGE recall scores. While Gigaword and DUC2004 contain abstractive ground-truth summaries, the remaining three datasets have token-level extractive ground-truth summaries. The ground-truth compressions in the Google sentence compression dataset \cite{filippova-altun-2013-overcoming} were automatically generated using grammatical constraints and distant supervision via headlines. The Broadcast and BNC datasets \cite{Clarke:Lapata:2008-ilp} contain manually created extractive sentence compressions which tend to be longer compared to the other evaluation datasets. Following previous work, we report a simple F1-score based on tokenized predicted and ground-truth summaries on the three extractive datasets, but also measure ROUGE F1 scores.

\begin{table}[]
\small
\centering
\begin{tabular}{l|l|l|l|l|l}
\textbf{Testset} & \textbf{Type} & \textbf{Size} & $L_{src}$ & $L_{tgt}$ & $cr$ \\
\hline
Gigaword         & abs           & 1951          & 29.7        & 8.8         & 0.4         \\
DUC2004          & abs           & 500           & 32.9        & 11.9        & 0.41        \\
Google           & ext           & 1000          & 27        & 11         & 0.45        \\
Broadcast        & ext           & 1370          & 19.8        & 14.4        & 0.76        \\
BNC              & ext           & 1629          & 27.9        & 19.3        & 0.72       
\end{tabular}
\caption{Overview of the evaluation datasets. The \textit{Type} column indicates whether the ground-truth is extractive or abstractive. \textit{Size} gives the number of sentences. }
\label{tab:datasets}
\end{table}

\subsection{Model Development}
\label{subsec:model-dev}

We tune our approach in several phases. At first, we identify an optimal learning rate and batch size using a grid search with a fixed training duration. We compare different settings based on the average reward achieved on a unlabelled, held-out set of the training data. Next, we test different values of $k$ (1, 5, 10, 50, 100), the number of samples per step, and pick the best $k$ based on the average reward on the validation set. This method of hyperparameter tuning is fully unsupervised.

Using learning rate $1e-05$, batch size $4$ and $k=100$ identified in the previous runs, we next compare the different options for the similarity reward listed in Section \ref{subsec:reward-functions} and pick the best (bi-encoder similarity) based on the F1-score on our labelled Newsroom-based validation set (see Appendix \ref{appendix:sim-reward}).

\subsection{Training}

We initialize the encoder component of our model with the pretrained 6-layer DistilRoBERTa model \citep{Sanh2019DistilBERTAD}. The binary classifier module is initialized randomly. We train each model for 8,000 steps with a batch size of 4 on a Google Cloud virtual machine with one NVIDIA Tesla T4 GPU, using the AdamW optimizer \citep{loshchilov2019decoupled}. Our default reward combination contains masked-LM fluency and bi-encoder similarity combined with either $R_{len}$ or $R_{cr}$. Table \ref{tab:models} gives an overview of the three models that are used in the evaluation. Note that the sample size of 100 is responsible for the long training durations. \textsc{SCRL-L8} and \textsc{SCRL-L11} are trained with $R_{len}$ whereas \textsc{SCRL-CR75} is trained with $R_{cr}$, with a compression ratio of $0.75$. This is because the ground-truth summary lengths are approximated better by a fixed length rather than a fixed ratio in the Google and DUC2004 datasets, whereas a fixed ratio describes the Broadcast and BNC datasets better. 

\begin{table}[]
\small
\centering
\begin{tabular}{l|l|l|l}
Name & Train data & Test Data & Time \\ \hline
SCRL-L8 & Gigaword & Gigaword & 9 \\
SCRL-L11 & Newsroom & DUC04, Google & 9.5 \\ 
SCRL-CR75 & Newsroom & Broadcast, BNC & 10
\end{tabular}
\caption{Overview of trained models and training time in hours.}
\label{tab:models}
\end{table}

\begin{table*}[t]
\small
\centering
\begin{tabular}{llcll|l|lll|l}
\hline
\multicolumn{1}{l|}{\textbf{Dataset}} & \multicolumn{1}{l|}{\textbf{Model}} & \multicolumn{3}{c|}{\textbf{ROUGE}} & \textbf{F1} & $L_{d}$ & $L_o$ & $cr_{o}$ & \textbf{Inf. Time (s)} \\ \hline
 \multicolumn{1}{l|}{} & \multicolumn{1}{l|}{} & \textbf{1} & \multicolumn{1}{c}{\textbf{2}} & \multicolumn{1}{c|}{\textbf{L}} &  &  &  & \multicolumn{1}{l|}{} & \textbf{} \\ \hline
\multicolumn{1}{l|}{} & \multicolumn{1}{l|}{Lead-L8} & \multicolumn{1}{l}{21.39} & 7.42 & \multicolumn{1}{l|}{20.03} & & 8 & 7.9 &  & \\
\multicolumn{1}{l|}{} & \multicolumn{1}{l|}{\citet{zhou2019simple}} & \multicolumn{1}{l}{26.48} & 10.05 & \multicolumn{1}{l|}{24.41}  &  &  & 9.3 &  & \\
\multicolumn{1}{l|}{\textbf{Gigaword}} & \multicolumn{1}{l|}{\citet{schumann2020discrete} L8} & \multicolumn{1}{l}{26.32} & 9.36 & \multicolumn{1}{l|}{24.19} & & 8 & 7.9 &  & \\
\multicolumn{1}{l|}{} & \multicolumn{1}{l|}{\citet{schumann2020discrete} L10} & \multicolumn{1}{l}{28.80} & \textbf{10.66} & \multicolumn{1}{l|}{25.82} & & 10 & 9.8 &  & \\
\multicolumn{1}{l|}{} & \multicolumn{1}{l|}{\textsc{HC}-L8} & \multicolumn{1}{l}{28.00} & 8.53 & \multicolumn{1}{l|}{25.90} & & 8 & 7.96 & 0.31 & 11.733 \\
\multicolumn{1}{l|}{} & \multicolumn{1}{l|}{\textsc{SCRL}-L8} & \multicolumn{1}{l}{\textbf{29.14}} & 9.98 & \multicolumn{1}{l|}{\textbf{26.57}}  & & 8 & 7.68 & 0.28 & \textbf{0.004} \\ \hline

\multicolumn{1}{l|}{\textbf{DUC2004}} & \multicolumn{1}{l|}{\citet{zajic2004bbn}\textsuperscript{$\bigstar$}} & \multicolumn{1}{l}{25.12} & 6.46 & \multicolumn{1}{l|}{20.12} &  &  &  &  & \\
\multicolumn{1}{l|}{} & \multicolumn{1}{l|}{\citet{baziotis2019seq3} } & \multicolumn{1}{l}{22.13} & 6.18 & \multicolumn{1}{l|}{19.3} &  &  &  &  & \\
\multicolumn{1}{l|}{} & \multicolumn{1}{l|}{\citet{west-etal-2019-bottlesum}} & \multicolumn{1}{l}{22.85} & 5.71 & \multicolumn{1}{l|}{19.87} &  &  &  &  & \\
\multicolumn{1}{l|}{} & \multicolumn{1}{l|}{\citet{schumann2020discrete} } & \multicolumn{1}{l}{\textbf{27.41}} & \textbf{8.76} & \multicolumn{1}{l|}{23.89} &  & 13 &  &  &  \\
\multicolumn{1}{l|}{} & \multicolumn{1}{l|}{\textsc{HC-L11}} & \multicolumn{1}{l}{27.40} & 8.65 & \multicolumn{1}{l|}{\textbf{24.16}} &  & 11 & 10.69 & 0.36 & 12.305 \\
\multicolumn{1}{l|}{} & \multicolumn{1}{l|}{\textsc{SCRL-L11}} & \multicolumn{1}{l}{25.27} & 7.82 & \multicolumn{1}{l|}{22.14} &  & 11 & 10.58 & 0.35 & \textbf{0.004} \\ \hline

\multicolumn{1}{l|}{} & \multicolumn{1}{l|}{Filipova\textsuperscript{$\bigstar$}} & & & & 0.82 &  &  & 0.38 &  \\
\multicolumn{1}{l|}{} & \multicolumn{1}{l|}{\citet{wang-etal-2017-syntax}\textsuperscript{$\bigstar$}} & & & & 0.8 &  &  & 0.43 & \\
\multicolumn{1}{l|}{} & \multicolumn{1}{l|}{\citet{wang2018sentence}} & & & & 0.565 &  &  & &  \\
\multicolumn{1}{l|}{\textbf{Google}} & \multicolumn{1}{l|}{\citet{zhou2019simple}} & & & & 0.61 &  &  &  &  \\
\multicolumn{1}{l|}{} & \multicolumn{1}{l|}{\citet{niu2019deleter}} & & & & 0.5 &  &  &  & \\
\multicolumn{1}{l|}{} & \multicolumn{1}{l|}{HC-L11} & 68.04 & 49.21 & 67.40 & 0.637 & 11 & 11.0 & 0.46 & 11.261 \\
\multicolumn{1}{l|}{} & \multicolumn{1}{l|}{SCRL-L11} & \textbf{70.22} & \textbf{53.03} & \textbf{69.84} & \textbf{0.711} & 11 & 10.8 & 0.44 &  \textbf{0.004} \\ \hline

\multicolumn{1}{l|}{} & \multicolumn{1}{l|}{\citet{wang-etal-2017-syntax}\textsuperscript{$\bigstar$}} & & & & 0.66 &  &  &  & \\
\multicolumn{1}{l|}{\textbf{Broadcast}} & \multicolumn{1}{l|}{\citet{wang2018sentence}} & & & & 0.665 &  &  &  &  \\
\multicolumn{1}{l|}{} & \multicolumn{1}{l|}{HC-CR75} & 82.20 & 63.78 & 81.76 & \textbf{0.792} & 75\% & 14.9 & 0.77 & 13.516 \\
\multicolumn{1}{l|}{} & \multicolumn{1}{l|}{SCRL-CR75} & \textbf{82.22} & \textbf{66.01} & \textbf{81.78} & 0.787 & 75\%  & 15.1 & 0.78 & \textbf{0.004} \\ \hline

\multicolumn{1}{l|}{} & \multicolumn{1}{l|}{\citet{wang-etal-2017-syntax}\textsuperscript{$\bigstar$}} & & & & 0.66 &  &  & 0.53 & \\
\multicolumn{1}{l|}{\textbf{BNC}} & \multicolumn{1}{l|}{\citet{wang2018sentence}} & & & & 0.675 &  &  &  &  \\
\multicolumn{1}{l|}{} & \multicolumn{1}{l|}{HC-CR75} & 78.91 & 60.10 & 78.13 & \textbf{0.768} & 75\% & 21.0 & 0.76 & 15.268 \\
\multicolumn{1}{l|}{} & \multicolumn{1}{l|}{SCRL-CR75} & \textbf{79.49} & \textbf{62.32} & \textbf{78.63} & 0.765 & 75\% & 21.0 & 0.76 & \textbf{0.004} \\ \hline
\end{tabular}
\caption{Results on evaluation datasets. \textsuperscript{$\bigstar$} indicates supervised models. ROUGE F1-scores are shown for all dataset but DUC2004, where ROUGE recall is used. $L_d$: desired output length, $L_o$ / $cr_o$: actual average length / compression ratio of the outputs.}
\label{tab:all-eval-results}
\end{table*}

\subsection{Baselines}
We compare our model to the greedy stochastic hill climbing approach in \citet{schumann2020discrete} which obtained state-of-the-art ROUGE results for unsupervised baselines on the Gigaword and DUC2004 datasets. Because this method and \textsc{SCRL} do not have identical objective functions, we implement the hill climbing algorithm applied to our reward functions, which we will name \textsc{HC} throughout this work. This allows for a clearer comparison between RL and discrete search. \textsc{HC} optimizes $R_{fluency}$, $R_{sim}$ under fixed length constraints instead of using $R_{len}$ and $R_{cr}$. Different from \citet{schumann2020discrete}, it runs for a fixed number of 2000 steps and restarts only when the search is stuck rather than in equal intervals (details in Appendix \ref{appendix:hc}). We analyze the performance of \textsc{HC} for different budgets to understand at what point search can surpass the learned policies. We also compare against \citet{zhou2019simple}, \citet{niu2019deleter} and the RL-based method by \citet{wang2018sentence} on datasets where results are available.  

\subsection{Evaluation Results}
\begin{figure}[h]
\centering
\includegraphics[width=1.0\columnwidth]{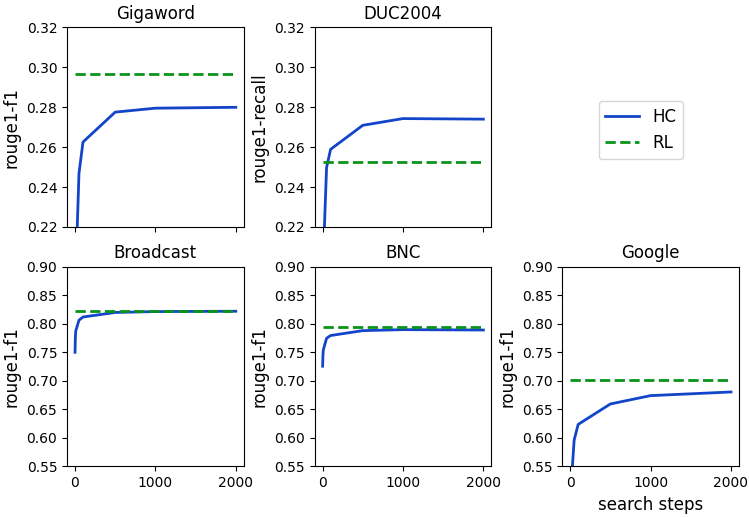}
\caption{Evaluation scores of RL  model compared to hill climbing algorithm (HC) at different search budgets.}
\label{fig:hc-vs-rl-eval}
\end{figure}

Table \ref{tab:all-eval-results} shows the evaluation results on all used test datasets. Results of methods apart from \textsc{SCRL} and \textsc{HC} are taken from previous works. We compute ROUGE scores using the implementation from Google Research\footnote{ \url{https://github.com/google-research/google-research/tree/master/rouge}}. On Gigaword, \textsc{SCRL} outperforms all baselines, except \citet{schumann2020discrete} with a 10 token constraint in ROUGE-2. On DUC2004, \textsc{SCRL} remains behind the hill climbing methods, but outperforms other unsupervised baselines. On the Google dataset, \textsc{SCRL} obtains state-of-the-art results among unsupervised methods. On Broadcast and BNC, \textsc{SCRL} and \textsc{HC} obtain very similar scores, which are both higher than previously reported results. Figure \ref{fig:hc-vs-rl-eval} shows ROUGE-1 scores obtained by \textsc{HC} at different search budgets, compared to \textsc{SCRL}. The hill climbing strategy approaches or outperforms the trained model at different paces, depending on the dataset.

Interestingly, \textsc{HC} still achieves higher rewards than \textsc{SCRL} relatively early during its search (see Appendix \ref{appendix:rewards}), which is inconsistent with the evaluation results. Potential reasons for this disparity are disadvantages through the hard length constraints, a mismatch between the heuristic reward functions and evaluation metrics, and beneficial biases induced through our training framework.

\subsection{Prediction Running Times}
\label{subsec:running-times}
We compare the inference-time speed of \textsc{SCRL} with \textsc{HC} using different budgets of search steps\footnote{On a Google Colab Notebook with a Tesla P-100 GPU}. The fastest batch size for both approaches is used. The \textit{Inference Time} in Table \ref{tab:all-eval-results} shows the average number of seconds per processed sentence, with the number of search steps set to $T=2000$ for \textsc{HC}. \textsc{SCRL} is roughly $4000 \times$ faster than \textsc{HC} with $T=2000$, and $\sim 200 \times$ faster when $T$ is reduced to $100$, for example. We believe that such a speed-up with a preserved evaluation performance is a critical factor when considering real-world applications of sentence compression.

\section{Analysis}
\label{sec:analysis}

\subsection{Summary Length and Extraction Regions}

\begin{figure}
    \includegraphics[width=1.0\columnwidth]{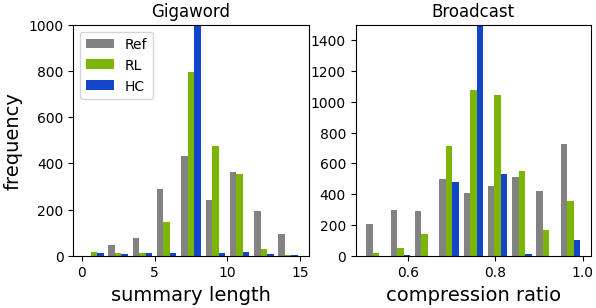}
    \caption{Distribution of summary lengths and compression ratios. The maximum frequencies of \textit{HC} are cropped.}
    \label{fig:summary-lengths}
\end{figure}

\begin{figure}
    \includegraphics[width=1.0\columnwidth]{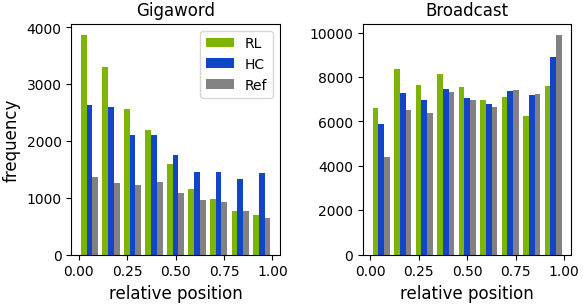}
    \caption{Distribution of relative positions from which tokens are extracted in source sentences.}
    \label{fig:positions}
\end{figure}

The length and compression ratio of summaries produced by \textsc{SCRL} is distributed around the desired values, with peakier distributions than in ground-truth summaries (examples in Figure \ref{fig:summary-lengths}). \textsc{HC} produces exactly the desired value whenever possible, due to the enforced constraint for length or ratio. Figure \ref{fig:positions} shows how \textsc{SCRL} and \textsc{HC} extract tokens from different relative positions within source sentences. \textsc{SCRL} has a higher tendency to extract early tokens. We hypothesize that this is a reliable high-reward strategy discovered during training, considering that a milder form of the lead-bias also shows in \textsc{HC}. Note that neither method is inherently biased in its design to prefer tokens from certain regions.

\begin{figure}
    \includegraphics[width=1.0\columnwidth]{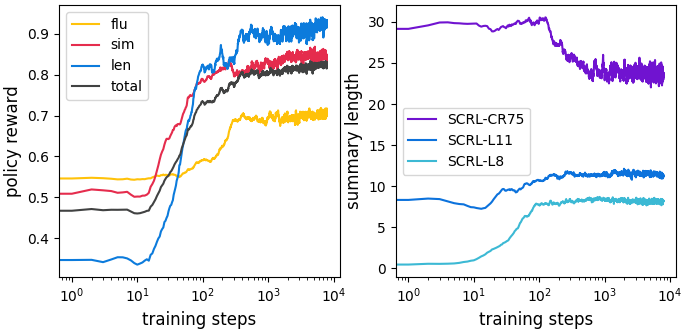}
    \caption{Development of reward functions (of SCRL-L11) and summary length during training (all models), from a moving average over 50 steps, using a log scale.}
    \label{fig:training-dynamics}
\end{figure}

\subsection{Training Dynamics}

Figure \ref{fig:training-dynamics} shows how rewards and summary length develop throughout training. The rewards generally increase quickly in the first few hundred training steps and then continue to grow very slowly. Fluency starts to increase later than the other reward functions, which is likely related to our observation that it is more sensitive to small changes in a summary. Interestingly, the summary lengths develop differently depending on the length or compression setting -- \textsc{SCRL-L8} and \textsc{SCRL-L11} start with short summaries and increase the  size over time whereas \textsc{SCRL-CR75} starts with long summaries before settling on a shorter certain range.

\begin{figure*}[t!]
\centering
\includegraphics[width=0.9\textwidth]{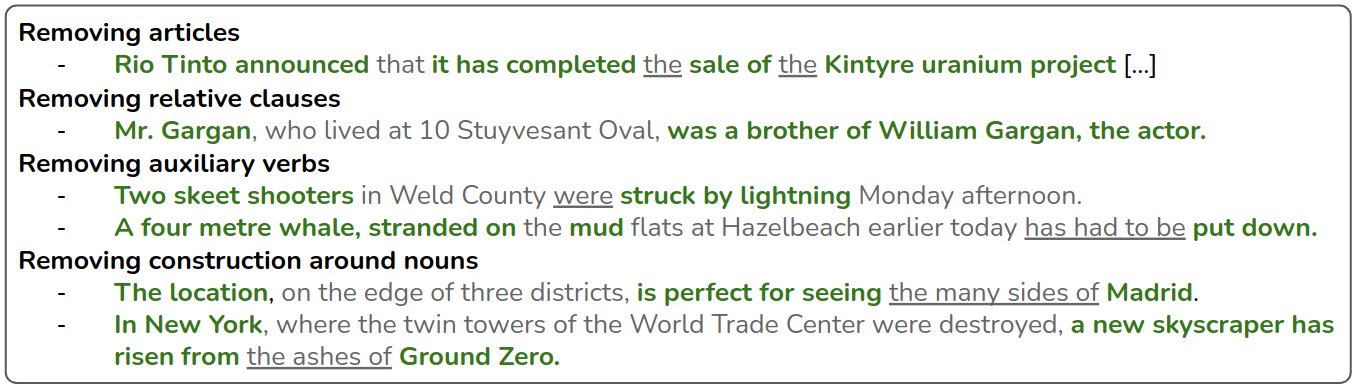}
\caption{Examples of learned summarization techniques. Selected tokens are marked in green and common removal behaviors are pointed out with underlining.}
\label{fig:learned-techniques}         
\end{figure*}

\subsection{Learned Summarization Techniques}
Our models learn a variety of behaviors to compress sentences, such as removing articles, auxiliary verbs, relative clauses and temporal expressions. Figure \ref{fig:learned-techniques} shows some examples.

\subsection{Error Analysis}

Even though our models learn to produce grammatical sentences fairly well, grammatical errors do still appear, and are more common for the models with a short output length (\textsc{SCRL-8}, \textsc{SCRL-11}). In some cases, semantic errors occur where the original meaning is changed or made unintelligeble. Both \textsc{SCRL} and \textsc{HC} are susceptible of semantic and grammatical errors, as can be seen in some examples in Appendix \ref{appendix:examples}. A type of error that is specific to \textsc{SCRL} is the splitting or merging of tokens resulting from its operation on Byte Pair Encoding-based subword tokens (more details in Appendix \ref{appendix:split-merge}).

\subsection{Customization via Reward Functions}

To demonstrate that our approach is flexible for customization, we pick a simple example of reprogramming model behavior using a hand-crafted reward function. We note that in some cases, the model unnecessarily keeps day references in compressed sentences, such as "Thursday" or "yesterday". We construct a simple reward function that returns zero if any day-like word from a small gazetteer appears in an output and a score of 1 otherwise. We fine-tune an existing model with this additional reward and observe that it successfully avoids including day-words that the previous model would include. Importantly, it additionally learned to remove other tokens attached to day-words, e.g. "on" in "on Monday", keeping the sentences grammatical. Table \ref{tab:customized} shows some examples. Empirically, the new model's outputs contain words from the gazeteer in 1\% of cases where they appear in the source, compared to 12\% in the initial model.

\begin{table}[h]
\small
\begin{tabular}{|l|l|}
\hline
\multicolumn{1}{|c|}{\textbf{Initial Model}} & \multicolumn{1}{c|}{\textbf{Customized Reward}} \\ \hline
\makecell{The burrito chain said \\ \textbf{on Tuesday} that comparable \\ sales fell 26.1\% last month.} & \makecell{The burrito chain said \\ that comparable sales \\ fell 26.1\%.} \\ \hline
\makecell{His car was found \\ \textbf{last Thursday}  \\ alongside Rubyvale Road.} & \makecell{His car was found \\ alongside \\ Rubyvale Road.} \\ \hline
\end{tabular}
\caption{Example outputs of model with customized reward function to exclude mentions of days.}
\label{tab:customized}
\end{table}

\vspace{-0.5cm}

\section{Discussion}
\label{sec:discussion}

We argue that RL offers the following advantages over discrete search strategies for sentence compression and similar text editing or generation tasks. The necessary search and exploration is moved into the training stage, allowing fast inference independently of how efficient objectives are to compute. Furthermore, discrete search unnecessarily spends time navigating through low-quality outputs that a trained model can quickly learn to avoid. Limitations of our approach compared to the search-based approach are its lesser flexibility in terms of on-the-fly customization and a sensitivity to disparities between training data and the application domain. Furthermore, the trained models show a lower capability to optimize the selected objectives compared to search, though this does not have a negative impact on the evaluation in most cases.

The fact that most of our training time is spent on estimating the quality of sampled compressions due to large sample size $k$, shows that our approach is somewhat similar to large-scale search strategies applied to a whole dataset, with the difference that the sampling behavior at each step changes over time and is informed by previous steps. This suggests that discrete search could support the RL training, similarly to the learning-from-search approach described by \cite{li2020unsupervised}.

\section{Conclusion}
\label{sec:conclusions}

This work presents a simple and effective approach for learning sentence compression models based on objective functions rather than ground-truth examples. Because it is unsupervised, it is well-suited for creating customized applications even when no gold training data is available, allowing for task-specific tuning based on arbitrary sets of reward functions, which do not need to be differentiable. Importantly, our approach is very fast at inference time compared to alternative discrete search-based methods. 
We are interested in several future directions related to this work: 1) systematic approaches to design reward functions for summarization, 2) RL-based summarization models with length control on the fly, 3) testing our approach on other languages, and 4) the design of curricula for different reward functions as they might pose varying difficulties at different stages of the training. 

\section*{Acknowledgments}
This work was funded by the Irish Research Council (IRC) under grant number EBPPG/2018/23, the Science Foundation Ireland (SFI) under grant number 12/RC/2289\_P2 and the enterprise partner Aylien Ltd. 
\bibliography{anthology,custom}
\bibliographystyle{acl_natbib}

\appendix

\section{Masked vs. Causal Language Model for Fluency}
\label{appendix:masked-vs-causal}

We compare the masked DistilRoBERTa\footnote{\url{https://huggingface.co/distilroberta-base}} language model to the causal DistilGPT2\footnote{\url{https://huggingface.co/distilgpt2}} model on our development dataset. Both models have roughly the same number of parameters (82M). Table \ref{tab:causal-lm} shows the results. 

\begin{table}[h]
\centering
\small
\begin{tabular}{@{}l|l|l@{}}
LM       & DistilRoBERTa & DistilGPT2 \\ \midrule
F1-Score &       0.565        & 0.546     
\end{tabular}
\caption{Comparison of a masked and a causal language model to estimate fluency.}
\label{tab:causal-lm}
\end{table}

\section{Similarity Functions}
\label{appendix:sim-reward}
Table \ref{tab:sim-comparison} compares different variants of the similarity reward functions on our development dataset.
\begin{table}[!h]
\centering
\small
\begin{tabular}{l|l|l|l}
Similarity & Bi-Sim & Cross-Sim & NLI \\ \hline
F1-Score & 0.624 & 0.598 & 0.564
\end{tabular}
\caption{Comparison of similarity reward functions on our development dataset.}
\label{tab:sim-comparison}
\end{table}

\section{Error Analysis: Split and Merged Tokens}
\label{appendix:split-merge}
One type of error is the splitting or merging of tokens from the source which results from the fact our model predicts labels at the level of BPE subword tokenization used in the pretrained language model that we finetune. While some of these occurrences are minor, e.g. 'St.' $\rightarrow$ 'St', or even useful compressions, e.g. '29th' $\rightarrow$ '29', many of these cases produce noisy outputs, e.g 'Perigord' $\rightarrow$ 'Perig'. Based upon analysis of prediction behavior, we estimate that $6\%$ of output sentences contain some form of this phenomenon.

\begin{algorithm}[!h]
    \small
    \caption{Stochastic First-Choice Hill Climbing}
    \label{algorithm:hc}
    \begin{algorithmic}
    \INPUT{objective~function~$R(x, y)$, source~sentence~$x$, summary~length~$L$, number~of~steps~$T$, initialization function $Init(x, L)$, sampling~function~$S(y)$}
    \STATE $y^0 \leftarrow Init(x, L)$
    \FOR{$t = 1$ to $T$}
        \STATE $y' \leftarrow S(y^{t-1})$
        \IF{${R(x, y')}\ge{R(x, {y}^{t-1})}$}
        \STATE{$y^t = y'$}
        \ELSE 
        \STATE{$ y^t = y^{t-1}$}
        \ENDIF
    \ENDFOR
    \RETURN $y_t$
\end{algorithmic}
\end{algorithm}

\section{Implementation Details}

\subsection{Pretrained Model IDs}

\begin{table*}[]
\centering
\small
\begin{tabularx}{\textwidth}{X|l}
\textbf{Model Usage} & \textbf{Model ID} \\ \hline
Encoder initialization for \textsc{SCRL} models & \texttt{distilroberta-base}  \\ \hline
Masked LM Fluency Reward & \texttt{distilroberta-base} \\ \hline
Causal LM Fluency Reward & \texttt{distilgpt2} \\ \hline
Bi-Encoder Similarity Reward & \texttt{all-distilroberta-v1}  \\ \hline
Cross-Encoder Similarity Reward & \texttt{cross-encoder/stsb-distilroberta-base} \\ \hline
Cross-Encoder NLI Reward & \texttt{cross-encoder/nli-distilroberta-base} \\ \hline
\end{tabularx}
\caption{Overview of pretrained models used throughout this work.}
\label{tab:models-ids}
\end{table*}

Table \ref{tab:models-ids} lists the model IDs of all open-source pretrained models used in this work, which can be found at \url{https://huggingface.co}.

\subsection{Tokenization}

We use the NLTK\footnote{\url{https://www.nltk.org/}} Punkt Tokenizer for several purposes in this work:

\begin{itemize}
    \item Obtaining the sentence length and compression ratio in $R_{len}$ and $R_{cr}$.
    \item Compressing sentences by selecting tokens in our hill climbing implementation \textsc{HC}.
    \item Obtaining source and summary tokens to compute the F1-score in Table \ref{tab:all-eval-results}, except for source and reference tokens on the Gigaword, Broadcast and BNC datasets which are pretokenized.
\end{itemize}

The involved transformer models (\textsc{SCRL}, $R_{fluency}$, $R_{sim}$) internally tokenize sentences based on Byte Pair Encoding.

\begin{figure*}[!b]
\centering
\includegraphics[width=0.9\textwidth]{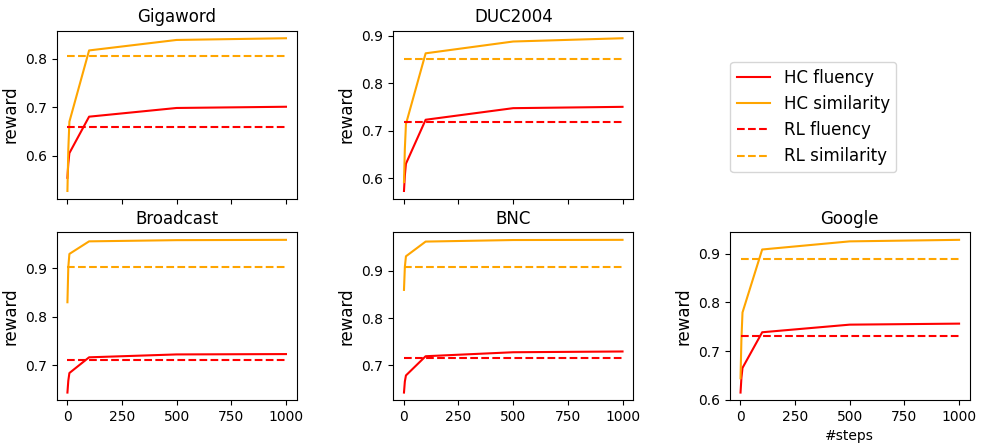}
\caption{Rewards of RL  model compared to hill climbing algorithm (HC) at different search budgets.}
\label{fig:hc-vs-rl-rewards}
\end{figure*}

\section{Hill Climbing Baseline}
\label{appendix:hc}

Algorithm \ref{algorithm:hc} shows the hill climbing search approach \textsc{HC} used in our experiments, which is based on \citet{schumann2020discrete}. At the beginning, a binary label sequence $y^0$ is initialized by setting $L$ randomly selected labels to $1$ and the rest to $0$. At each step $t$, $S(y)$ samples a new label sequence $y'$ by randomly selecting a positive and a negative-valued label $y_i$ and $y_j$ and swapping their value. Note that this always keeps the number of tokens at $L$. The sampled $y'$ is accepted if it obtains a higher or equal objective score than the previously best candidate $y^{t-1}$. We keep track of previously created sequences and skip these. If no new label sequence can be discovered at step $t$, we terminate the algorithm and restart it with $T-t$ remaining steps. In the end, the highest-scoring $y$ found across different runs is returned.
We generally set $T$ to $2000$ and keep track of intermediate results to evaluate \textsc{HC} also at fewer search steps. Due to this, we decided restart the search dynamically rather than in equal-paced intervals, which we believe should be tuned with respect to a known maximum budget $T$.

\section{Rewards Obtained by \textsc{HC}  vs. \textsc{SCRL}}
Figure \ref{fig:hc-vs-rl-rewards} compares the $R_{fluency}$ and $R_{sim}$ rewards of \textsc{SCRL} to  \textsc{HC} with different search budgets. The length and compression ratio rewards are not included as these are enforced through a constraint by \textsc{HC}. Note that these figures need to be interpreted carefully as they assume that both approaches produce summaries of comparable lengths. For example, the similarity reward tends to increase with the summary length.

\label{appendix:rewards}

\section{Output Examples}
\label{appendix:examples}
Table \ref{tab:output-examples} lists a few examples outputs produced by \textsc{SCRL} and \textsc{HC}.

\begin{table*}[]
\centering
\begin{tabularx}{\textwidth}{l|X}
 \hline \hline
 \textbf{Source} & the us space shuttle atlantis separated from the orbiting russian mir space station early saturday , after three days of test runs for life in a future space facility , nasa announced . \\ \hline
\textbf{SCRL-L8} & the space shuttle atlantis separated from russian station \\ \hline
\textbf{HC-L8} & atlantis space station after test runs for nasa \\ \hline \hline

\textbf{Source} & a katyusha rocket fired from lebanon on saturday morning hit the western galilee in north israel , causing two lightly hurt , israel radio reported . \\ \hline
\textbf{SCRL-L8} & katyusha rocket fired from lebanon hit galilee israel \\ \hline
\textbf{HC-L8} & katyusha rocket fired hit western galilee \colorbox{pink}{israel israel}  \\ \hline \hline

\textbf{Source} & Manchester United have agreed a £35m deal to sign Sporting Lisbon midfielder William Carvalho, according to talkSPORT. \\ \hline
\textbf{SCRL-L11} & Manchester United agreed £35m deal to sign Lisbon midfielder William Carvalho. \\ \hline
\textbf{HC-L11} & Manchester United have agreed a £35m deal to sign William Carvalho \\ \hline \hline

\textbf{Source} & Egyptian President Hosni Mubarak met here Sunday with Syrian President Hafez Assad to try to defuse growing tension between Syria and Turkey. \\ \hline
\textbf{SRCL-L11} & Egyptian President Hosni Mubarak met with Syrian President Hafez Assad \colorbox{pink}{def.} \\ \hline
\textbf{HC-L11} & Egyptian President Hosni Mubarak met Sunday with Syrian President Hafez Assad \\ \hline \hline

\textbf{Source} & Russian President Boris Yeltsin, who is still recuperating from his latest illness, has canceled a trip to an Asian summit next month, his office said Friday. \\ \hline
\textbf{SCRL-L11} & Russian President Boris Yeltsin recuperating has canceled a trip to Asian summit. \\ \hline
\textbf{HC-L11} & Russian President Boris Yeltsin has canceled trip to an Asian summit \\ \hline \hline

\textbf{Source} & Laurie had a passion and a warmth for people rather than the state . \\ \hline
\textbf{SCRL-CR75} & Laurie had a passion and warmth for people. \\ \hline
\textbf{HC-CR75} & Laurie had a passion and \colorbox{pink}{warmth for the state} . \\ \hline \hline

\textbf{Source} & And speaking of the royals , the Duchess of York , Sarah Ferguson , was in Los Angeles last week holed up at the Four Seasons Hotel and when she ventured out , I hear she visited some of the studios like Sony to have meetings involving TV projects . \\ \hline
\textbf{SCRL-CR75} & And speaking of the royals, the Duchess of York, Sarah Ferguson, was in Los Angeles last week holed up at the Four Seasons Hotel and I hear she visited studios like Sony to have meetings. \\ \hline
\textbf{HC-CR75} & And speaking of royals , Duchess of York Sarah Ferguson was in Los Angeles last week at the Four Seasons Hotel and when she ventured out she visited some of the studios like Sony to have meetings . \\ \hline \hline

\textbf{Source} & Of the 24,058 people interviewed , 37.7 per cent of women attended arts events and 33.1 per cent of men . \\ \hline
\textbf{SCRL-CR75} & Of 24,058 people interviewed, 37.7 per cent of women attended arts events \colorbox{pink}{and 33.1.} \\ \hline
\textbf{HC-CR75} & Of 24,058 people interviewed , 37.7 per cent women attended arts events and 33.1 men . \\ \hline \hline
\end{tabularx}
\caption{Output examples, with semantic and grammatical errors highlighted.}
\label{tab:output-examples}
\end{table*}

\end{document}